%% file: bare_jrnl.tex
\documentclass[journal]{IEEEtran}
\usepackage{graphicx}
\usepackage[ruled,vlined]{algorithm2e}
\usepackage{multirow}
\usepackage{makecell}
\usepackage{comment}
\usepackage{amsmath,amssymb} 
\usepackage{color,colortbl}
\definecolor{LightCyan}{rgb}{0.88,1,1}

\newcommand{\eat}[1]{}

\newcommand{\abovesec}{0pt}
\newcommand{\belowsec}{0pt}

\def\eg{\emph{e.g}.} 
\def\ie{\emph{i.e}.} 
\def\cf{\emph{cf}.} 
\def\etc{\emph{etc}.} 
 
\def\etal{\emph{et al}.}

\ifCLASSINFOpdf
\else
\fi
\begin{document}
%
\title{Grounding-Tracking-Integration}
%
%
%
\author{Zhengyuan~Yang,~\IEEEmembership{Student Member,~IEEE,}
        Tushar~Kumar, 
        Tianlang~Chen,~\IEEEmembership{Student Member,~IEEE,}\\
        Jingsong~Su, 
        and~Jiebo~Luo,~\IEEEmembership{Fellow,~IEEE}
\thanks{Z. Yang, T. Kumar, T. Chen and J. Luo are with the Department of Computer Science, University of Rochester, Rochester, NY, 14627 USA (e-mail: \{zyang39, tusharku, tchen45, jluo\}@cs.rochester.edu).}
\thanks{J. Su is with Xiamen University, Xiamen, China (e-mail: jssu@xmu.edu.cn).}
\thanks{Corresponding author: Jiebo Luo.}
\thanks{Copyright \textcopyright 2020 IEEE. Personal use of this material is permitted. However, permission to use this material for any other purposes must be obtained from the IEEE by sending an email to pubs-permissions@ieee.org.}
}

\maketitle

\begin{abstract}
    \input{abs}
\end{abstract}


%
\IEEEpeerreviewmaketitle

\input{intro}
\input{related}
\input{approach_new}
\input{exp_new}
\input{conclusion}

\section*{Acknowledgment}
This work is supported in part by NSF awards IIS-1704337, IIS-1722847, and IIS-1813709, Twitch Fellowship, as well as our corporate sponsors.


\ifCLASSOPTIONcaptionsoff
  \newpage
\fi

\bibliographystyle{IEEEtran}
\bibliography{ref.bib}



%



\begin{IEEEbiography}[{\includegraphics[width=1in,height=1.1in,clip]{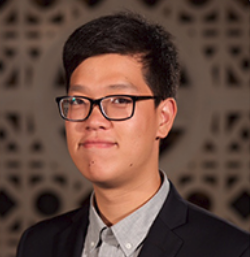}}]{Zhengyuan Yang}
received the BE degree in electrical engineering from the University of Science and Technology of China in 2016. He is a Ph.D. candidate in Computer Science at University of Rochester, Rochester, NY, advised by Prof. Jiebo Luo. His research interests mainly include vision+language and multimodal learning.
\end{IEEEbiography}

\begin{IEEEbiography}
[{\includegraphics[width=1in,height=1.1in,clip]{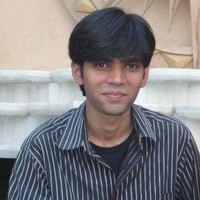}}]{Tushar Kumar} received the BE degree in information technology from Delhi College of Engineering in 2013, and the Master's degree in computer sciences from the University of Rochester in 2020.
\end{IEEEbiography}

\begin{IEEEbiography}
[{\includegraphics[width=1in,height=1.25in,clip]{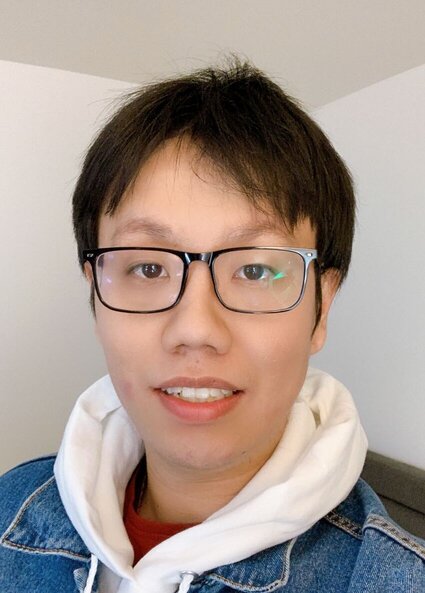}}]{Tianlang Chen} received the BE degree in electrical engineering from the University of Science and Technology of China in 2016. He is currently pursuing the PhD degree with the Computer Science Department, University of Rochester, under the supervision of Prof. Jiebo Luo. His research interests mainly include tasks related to joint visual-textual learning such image captioning, image-text matching, visual question answering and visual grounding.
\end{IEEEbiography}

\begin{IEEEbiography}
[{\includegraphics[width=1in,height=1.25in,clip]{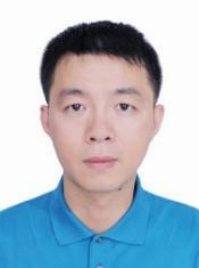}}]{Jinsong Su}  received his PhD degree in computer science at the Institute of Computing Technology of the Chinese Academy of Sciences in July, 2011. He is now an associate professor of Software School in Xiamen University. His research interest is natural language processing and neural machine translation. 
\end{IEEEbiography}

\begin{IEEEbiography}
[{\includegraphics[width=1in,height=1.25in,clip]{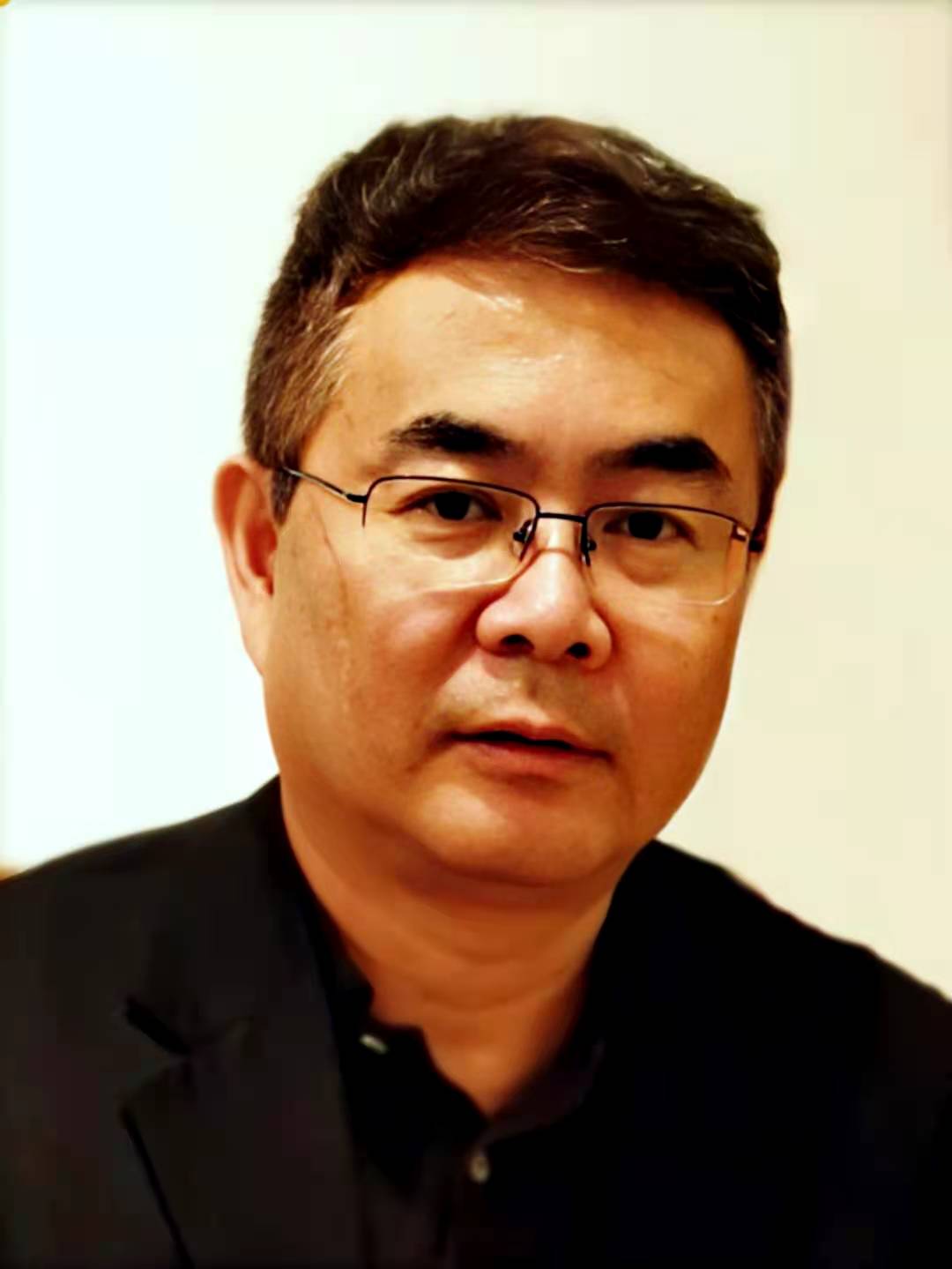}}]{Jiebo Luo} (S93, M96, SM99, F09) is a Professor of Computer Science at the University of Rochester which he joined in 2011 after a prolific career of fifteen years at Kodak Research Laboratories. He has authored over 400 technical papers and holds over 90 U.S. patents. His research interests include computer vision, NLP, machine learning, data mining, computational social science, and digital health. He has been involved in numerous technical conferences, including serving as a program co-chair of ACM Multimedia 2010, IEEE CVPR 2012, ACM ICMR 2016, and IEEE ICIP 2017, as well as a general co-chair of ACM Multimedia 2018. He has served on the editorial boards of the IEEE Transactions on Pattern Analysis and Machine Intelligence (TPAMI), IEEE Transactions on Multimedia (TMM), IEEE Transactions on Circuits and Systems for Video Technology (TCSVT), IEEE Transactions on Big Data (TBD), ACM Transactions on Intelligent Systems and Technology (TIST), Pattern Recognition, Knowledge and Information Systems (KAIS), Machine Vision and Applications, and Journal of Electronic Imaging. He is the current Editor-in-Chief of the IEEE Transactions on Multimedia. Professor Luo is also a Fellow of ACM, AAAI, SPIE, and IAPR. 

\end{IEEEbiography}

\end{document}

%% file: abs.tex
In this paper, we study \textbf{tracking by language} that localizes the target box sequence in a video based on a language query. We propose a framework called GTI that decomposes the problem into three sub-tasks: \textbf{G}rounding, \textbf{T}racking, and \textbf{I}ntegration. The three sub-task modules operate simultaneously and predict the box sequence frame-by-frame. ``Grounding'' predicts the referred region directly from the language query. ``Tracking'' localizes the target based on the history of the grounded regions in previous frames. ``Integration'' generates final predictions by synergistically combining grounding and tracking. With the ``integration'' task as the key, we explore how to indicate the quality of the grounded regions in each frame and achieve the desired mutually beneficial combination. To this end, we propose an ``RT-integration'' method that defines and predicts two scores to guide the integration: 1) R-score represents the Region correctness whether the grounding prediction accurately covers the target, and 2) T-score represents the Template quality whether the region provides informative visual cues to improve tracking in future frames. We present our real-time GTI implementation with the proposed RT-integration, and benchmark the framework on LaSOT and Lingual OTB99 with highly promising results. Moreover, we produce a disambiguated version of LaSOT queries to facilitate future tracking by language studies.

%% file: intro.tex
\section{Introduction}

\begin{figure}[t]
\begin{center}
   \centerline{\includegraphics[width=8.0cm]{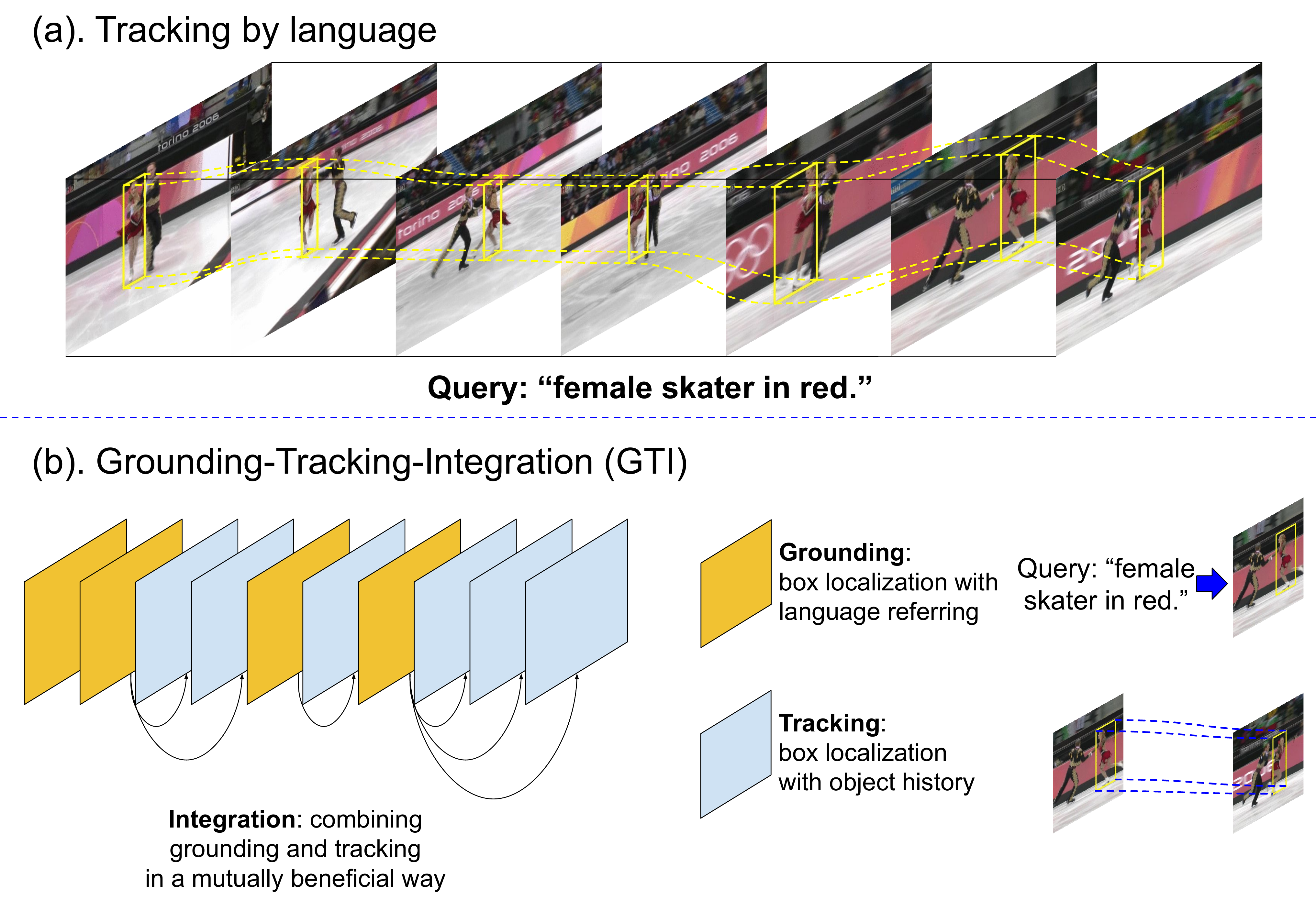}}
\end{center}
\vspace{-0.2in}
    \caption{\textit{Tracking by language} aims to localize the tubelet specified by a language query. We propose a GTI framework that decomposes the problem into three sub-tasks: grounding, tracking, integration. This study focuses on the key ``integration'' task.
    }
\label{fig:intro}
\end{figure}

\IEEEPARstart{G}{iven} a video and a language query, tracking by language~\cite{li2017tracking} is the task of predicting the box sequence of the referred object based on the input language query, as shown in Figure~\ref{fig:intro} (a). The grounded box sequences are predicted sequentially in each frame of the input video. Compared to specifying the target by drawing a box as in object tracking~\cite{VOT_TPAMI,wu2013online,wu2015object,yilmaz2006object}, providing a language query is a natural way of human-computer interaction. The language specification provides the clear semantic meaning of the target and thus alleviates certain failures in object tracking caused by appearance changes, occlusion, box drifting, \etc~ Tracking by language also opens up applications such as starting at an arbitrary time-step and searching in a video corpus in parallel. In addition, good tracking by language model benefits various related research problems, such as language-based video retrieval~\cite{yamaguchi2017spatio} and video QA~\cite{lei2018tvqa}.

Naturally, two kinds of information are available in tracking by language. On the one hand, the language query contains target specifications in all frames. On the other hand, the history of the grounded image patches in previous frames provides cues for the target. Therefore, tracking by language can be approached either from language referring (``grounding'') or visual patch matching (``tracking'') perspectives. For the first perspective, ``grounding'' approaches the problem by processing each frame independently. However, ``grounding'' methods frequently fail in frames of degraded visual qualities\eat{ such as the one with motion blurry, illumination changes, etc}. The grounded regions also tend to be inconsistent throughout time, as ``grounding'' alone exploits no neighboring frame similarities in videos. For the second perspective, ``tracking'' localizes the region based on a given box in previous frames. When initialized with an ideal given box (by grounding), ``tracking'' generally provides tubelets of better qualities than ``grounding''. However, `tracking'' suffers from bad initialization when the language grounded region either refers to the incorrect object or does not contain informative visual cues of the target for tracking.

This study builds on the understanding that neither ``grounding'' nor ``tracking'' alone solves the tracking by language problem, while the combination can compensate for each other's weaknesses. ``Tracking'' has the potential to correct ``grounding'' failures based on the information from adjacent frames, whereas ``grounding'' could improve ``tracking'' by re-initializing the tracker with better language grounded regions.

In this study, we propose a \textbf{GTI} framework, where we decompose the tracking by language task into three sub-tasks: \textbf{G}rounding, \textbf{T}racking and \textbf{I}ntegration. Given a frame, ``grounding'' localizes the region directly from the input language query. ``Tracking'' predicts by using the history of grounded regions as tracking templates, \ie, the ``tracking'' predicted region should be visually similar to the region in tracking templates. ``Integration'' combines the two perspectives in a mutually beneficial way to obtain better final predictions. As shown in Figure~\ref{fig:intro} (b), ``integration'' selects whether ``grounding'' or ``tracking'' \eat{should be focused more on} is more important in each frame, and generates the final box prediction accordingly. In frames where ``grounding'' is assigned higher importance, the language grounded region is included in the region history to help ``tracking'' in future frames. The three modules function simultaneously to generate tubelet predictions frame-by-frame.

\noindent {\bf Criteria for integration.}
While a wide range of ``grounding''~\cite{plummer2017flickr30k,yang2019fast,yu2018mattnet,yu2016modeling} and ``tracking''~\cite{danelljan2017eco,li2019siamrpn++,li2018high} methods exist, the ``integration'' problem is unique in tracking by language and we are not aware of any proper method that can be directly applied. ``Integration'' with pre-defined rules or fixed weights in all frames~\cite{li2017tracking} generally shows limited performance. Because such naive methods operate independently of the per-frame context and grounded regions, they neither manage to correct grounding failures with tracking results nor strengthen future tracking with selected grounded regions. Instead, the ``integration'' module \eat{should adaptively adjust the grounding-tracking importance} should operate adaptively in each frame by referencing the corresponding visual input, language query, and grounded region. To be specific, a good ``integration'' module should satisfy the following criteria: 1) The module should predict if the grounded region accurately covers the target, and assign higher importance to ``grounding'' in such frames. 2) The module should predict if the grounded region contains informative visual cues of the target that could improve the tracker, and include such region into the object history. 3) The module should be light-weighted and fast.

\noindent {\bf Mechanism for integration.}
We propose a new paradigm for the ``integration'' problem named \textit{RT-integration}. In each frame, we predict two scores to guide the ``integration''. R-score reflects the \textbf{R}egion correctness, \ie, whether the grounded region accurately covers the language referred object. T-score reflects the \textbf{T}emplate quality, \ie, whether the grounded region contains discriminative visual cues to help ``tracking''. High RT-scores indicate the high importance of ``grounding''. In such frames, we take ``grounding'' predictions both as the outputs and future tracking templates, whereas in the remaining frames, the ``tracking'' prediction is adopted as the outputs to correct possible grounding failures. We derive the ground-truth RT-scores from box annotations and train a separate module for RT-score prediction.

Finally, we present our real-time implementation of the GTI framework with the proposed RT-integration. We benchmark the proposed framework on LaSOT~\cite{fan2019lasot} and Lingual OTB99~\cite{li2017tracking} with highly promising results. As the original language queries in LaSOT can be ambiguous~\cite{fan2019lasot}, we clean the dataset by replacing the ambiguous queries with new annotations. Our contributions are:
\begin{itemize} 
\item We propose a Grounding-Tracking-Integration (GTI) framework for tracking by language.
\item We propose ``RT-integration'' that adaptively integrates grounding and tracking with the region correctness score and template quality score predicted in each frame.
\item Our real-time implementation of the GTI framework shows highly promising results on multiple datasets.
\item We clean up the ambiguous queries in LaSOT~\cite{fan2019lasot} to facilitate future tracking by language studies. 
\end{itemize}

%% file: related.tex
\section{Related Work}
\noindent {\bf Tracking with box specifications.}
Tracking returns the tubelet of the specified object in a video. Our study is related to the traditional object tracking~\cite{VOT_TPAMI,wu2013online,wu2015object,yilmaz2006object}, where the ground-truth box in the first frame (the tracking template) specifies the object of interest. Correlation filter based methods~\cite{danelljan2015learning,danelljan2014adaptive,henriques2014high} show good efficiency and accuracy on the task. Recently, the Siamese network based trackers~\cite{bertinetto2016fully,li2019siamrpn++,li2018high} also show promising performance. In this study, we study the problem of using language queries to replace boxes as the target specification.

\noindent {\bf Tracking with language cues.}
Several previous studies explore tracking with language cues~\cite{fan2019lasot,li2017tracking,shi2019not,wang2018describe}.
Wang et al.~\cite{wang2018describe} adopt language queries as the extra information alongside with boxes for tracking. LaSOT~\cite{fan2019lasot} is a recently proposed large scale tracking dataset that has auxiliary language query annotations. Li et al.~\cite{li2017tracking} first introduce the tracking by language task and propose a Lingual Specification Attention Network (LSAN). The authors encode the region history and language query as the parameters for two independent dynamic filters, and generate per-frame tracking and grounding predictions accordingly. The predictions are then fused with a fixed weight in all frames. We later show that LSAN is a special case of the GTI framework with a naive integration module. Feng~\etal~\cite{feng2020real} propose to solve tracking by language with the tracking by detection approach, and the tracking prediction is fed into the detectors to refine the detection results.

\noindent {\bf Visual grounding.}
Visual grounding~\cite{kazemzadeh2014referitgame,plummer2017flickr30k,yu2016modeling} is the task of localizing the referred region in an image given a language query. Most previous methods~\cite{plummer2018conditional,sadhu2019zero,wang2019learning,yu2018mattnet,yu2017joint} follow a two-stage pipeline, where a number of region candidates are first detected, followed by a language-based ranking stage to find the most relevant region. The recently proposed one-stage methods~\cite{chen2018real,yang2019fast,yang2020improving} conduct visual-textual fusion at image level and improve both the accuracy and inference speed. Recent studies~\cite{zhang2020does,sadhu2020video} explore the visual grounding problem in videos.

\noindent {\bf Self-evaluation scores.}
Our proposed ``RT-integration'' module is related to previous studies~\cite{danelljan2019atom,jiang2018acquisition,goldman2019precise,huang2019mask} on learning self-evaluation scores. In object tracking, ATOM~\cite{danelljan2019atom} predicts the Intersection over Union (IoU) between the tracking output and the ground-truth target by taking tracking templates as references. Our method is more relevant to IoU prediction in object detection~\cite{jiang2018acquisition} and instance segmentation~\cite{huang2019mask}, where no template references are available. IoU-Net~\cite{jiang2018acquisition} proposes an IoU prediction module on top of the detection backbone~\cite{lin2017feature} to predict the localization confidence. MS R-CNN~\cite{huang2019mask} extends the idea for instance segmentation. 

Previous studies in video object detection~\cite{kang2016object,tang2019object} explore the similar idea of score-based integration. Tang~\etal~\cite{tang2019object} propose to generate accurate and reliable object tubelet prediction by linking short tubelets based on the temporal overlap. Kang~\etal~\cite{kang2016object} show better video object detection can be achieved by combining the confidence score of a per-frame object detector and a tracker.
Going beyond self-evaluation score prediction based on the objectiveness, the ``integration'' task in tracking by language poses extra requirements of 1) predicting visual-textual similarity, 2) predicting template quality, and 3) being fast.

%% file: approach_new.tex
\section{Grounding-Tracking-Integration}
\begin{figure}[t]
\begin{center}
   \centerline{\includegraphics[width=7cm]{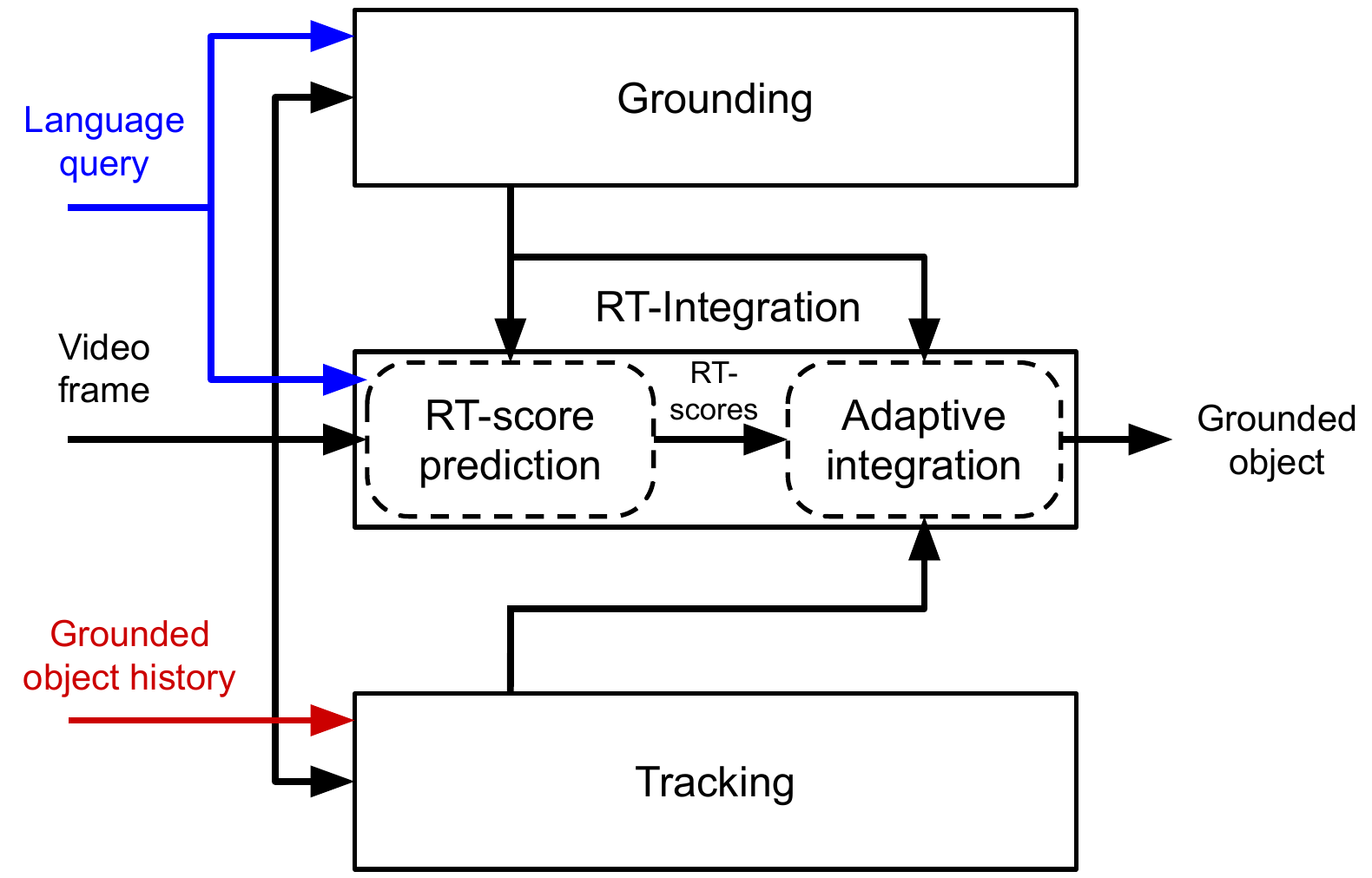}}
\end{center}
	\caption{The block diagram of the GTI framework with our proposed RT-integration.}
\label{fig:gti}
\end{figure}

Given a natural language query for a video, we hope to return the box sequence of the referred object. Different from object tracking~\cite{li2019siamrpn++,li2018high}, the references are specified by a language query instead of the ground-truth bounding box in the first frame. We propose a Grounding-Tracking-Integration (GTI) framework to approach the problem. As shown in Figure~\ref{fig:gti}, the three modules operate simultaneously and generate box predictions frame-by-frame. In each frame, ``grounding'' takes the frame and language query as input for object localization. ``Grounding'' operates independently in each frame and does not accumulate errors. However, it may fail due to the errors of grounding methods.
``Tracking'' predicts the box based on the history of language grounded regions. When provided a correct region in nearby frames as the template, ``tracking'' generally generates better box predictions than ``grounding.'' However, ``tracking'' often accumulates the error from the given template, and decreases in performance when the temporal distance between the template and current frame increases. ``Integration'' looks at both grounding and tracking predictions, and generates the final prediction.

\vspace{\abovesec}
\section{RT-Integration}
\vspace{\belowsec}
\label{sec:4}

We investigate the ``integration'' task in the GTI framework. The goal of this sub-task is to combine ``grounding'' and ``tracking'' in a mutually beneficial and overall synergistic way to generate better final predictions. In frames where ``grounding'' \eat{likely succeeds} predictions are of good quality, including such grounded regions into tracking templates strengthens the tracker for future frames. In  frames where ``grounding'' is likely to fail, adopting the ``tracking'' prediction generally leads to better final predictions. To achieve such a mutually beneficial combination, ``integration'' should predict when ``grounding'' is of good quality or likely to fail, and adjust the grounding-tracking importance in each frame accordingly.

The core idea of our proposed RT-integration is to represent the grounding and tracking importance in each frame as two scores, namely the RT-scores, where higher score values indicate the better quality and thus higher importance of ``grounding.'' The R-score reflects if the grounded region precisely covers the target, and the T-score shows if the grounded region contains visual cues that can improve the tracker. 
In frames with high RT-scores, the ``grounding'' prediction is selected as the output and used to update the tracker, whereas the remaining frames are processed by ``tracking.'' This study focuses on how to properly define and precisely predict the RT-scores.

\begin{figure*}[t]
\begin{center}
   \centerline{\includegraphics[width=14cm]{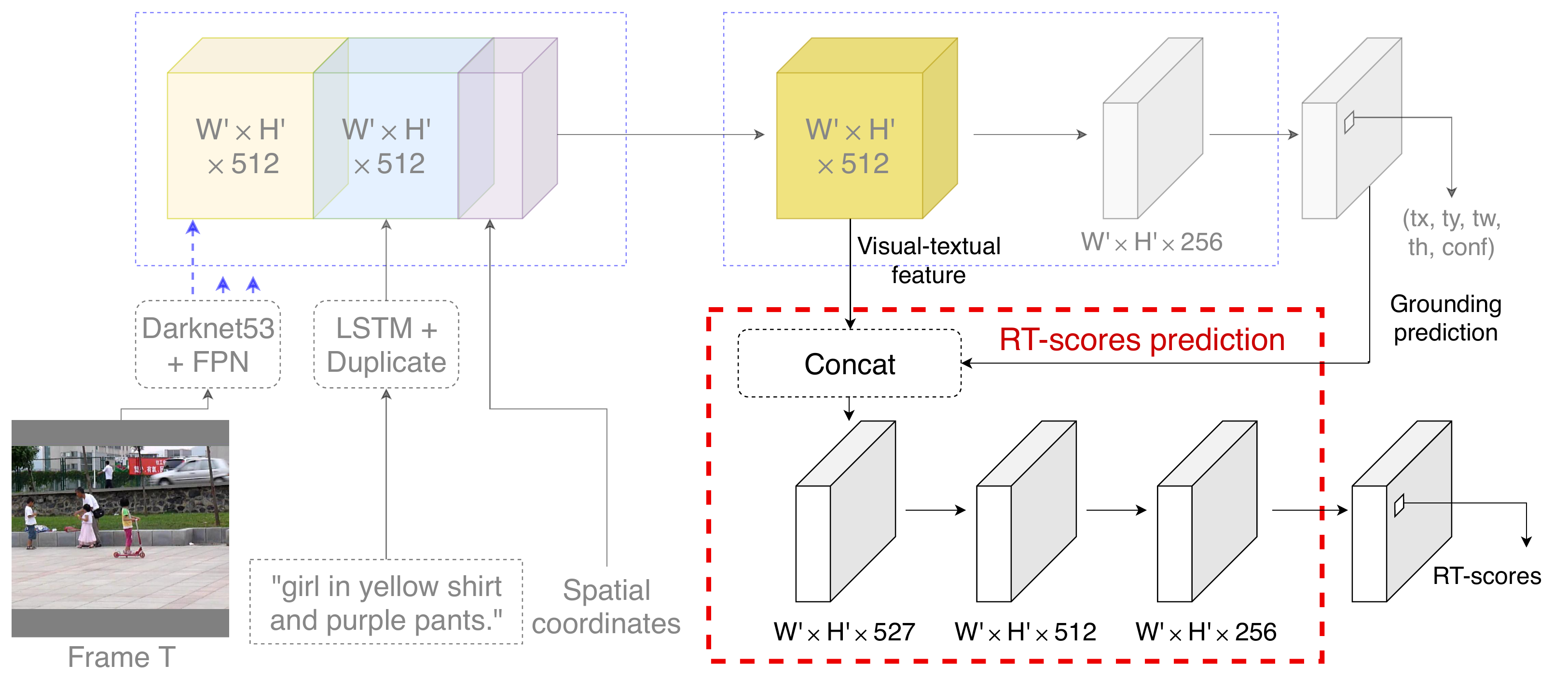}}
\end{center}
\vspace{-0.2in}
	\caption{The network architecture for RT-scores prediction. The backbone grounding method~\cite{yang2019fast} is shown in translucent colors outside the red box. Feature pyramid heads are used in the grounding method. For visualization purpose, we only show one head.}
\label{fig:arch}
\end{figure*}

A separate module is trained in a fully supervised way to predict the RT-scores, as shown in Figure~\ref{fig:gti}. The input to the module is the visual-textual feature and the language grounded region in a frame. The output is the corresponding RT-score prediction, as shown in Figure~\ref{fig:arch}. During training, the ground-truth RT-scores are derived from the box annotations and are used to train the module. Essentially, ``integration'' can be regarded as a self-judge process for the framework to examine whether the language grounded region in a frame is valid as the output and new template.
Section~\ref{sec:score} introduces the definition of the derived RT-scores. Section~\ref{sec:arch} presents the details of the module architecture and training procedure. During inference, RT-scores are predicted with the trained module in each frame, and guide the adaptive integration that synergistically combines grounding and tracking to generate final predictions.
\eat{combines grounding and tracking predictions as the final output. The grounding, tracking and integration modules operate simultaneously and generate the tubelet prediction frame-by-frame.}
Section~\ref{sec:int} introduces the RT-score-guided adaptive integration in each frame. A complete inference pipeline of the GTI framework is presented in Section~\ref{sec:imp}.

\begin{figure}[t]
\begin{center}
   \centerline{\includegraphics[width=8.5cm]{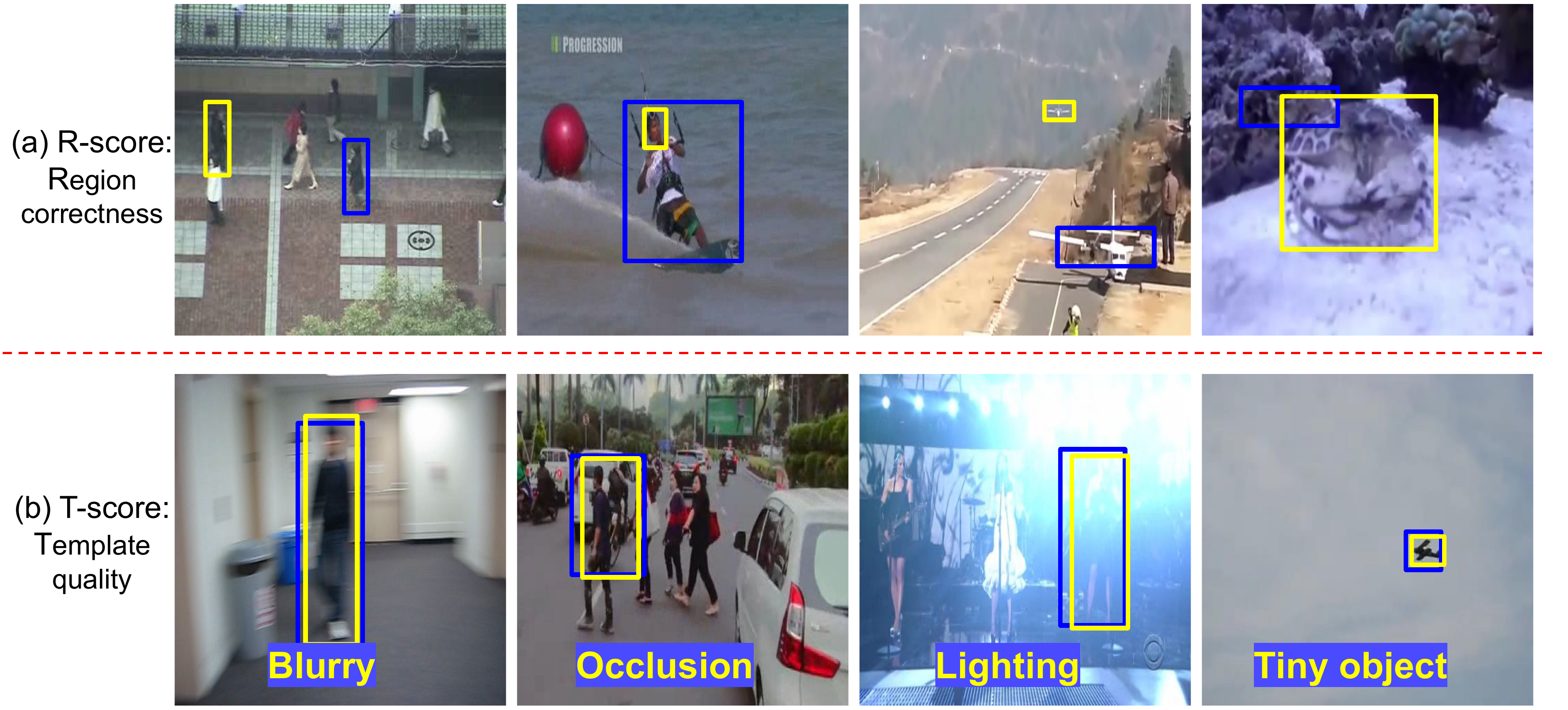}}
\end{center}
\vspace{-0.1in}
	\caption{Example frames with low region correctness scores (top row) and low template quality scores (bottom row). Blue/ yellow boxes are grounding predictions~\cite{yang2019fast}/ ground-truth, respectively.}
\label{fig:temp}
\end{figure}
\vspace{\abovesec}
\subsection{RT-scores}
\vspace{\belowsec}
\label{sec:score}

Two factors are essential for an ideal ``integration.'' First, ``integration'' should predict if the language grounded region accurately covers the target. The state-of-the-art grounding method~\cite{yang2019fast} commonly fails in frames with multiple objects of the same kind, tiny targets, and limited visual qualities (\eg, the top row in Figure~\ref{fig:temp}), when ``tracking'' should be adopted to correct the errors. Second, ``integration'' should predict if the grounded region contains visual cues that can strengthen the tracker to help future frames. The bottom row in Figure~\ref{fig:temp} shows negative examples with limited frame qualities or improper target statuses.

We propose to model the two factors with two scores respectively, namely the RT-scores. The R-score (\textbf{R}egion correctness) models how accurately the grounded box covers the target. In frames with low R-scores, grounding is likely to be failed and can be corrected by tracking. We define the R-score as the Intersection over Union (IoU) between the language grounded region and the ground-truth box. We collect the per-frame visual-textual feature and grounded region pairs with a visual grounding method~\cite{yang2019fast} and calculate the R-score in each sample accordingly.
The T-score (\textbf{T}emplate quality) models how well the target image patch in a frame serves as the tracking template, \ie, if purely relying on the visual similarity between the target image patch (the tracking template) and the candidates in future frames, how accurate the localization results will be. In frames with high T-scores, the grounding predictions contain informative visual cues that could improve tracking, while some patches have low T-scores and do not benefit tracking (as shown in Figure~\ref{fig:temp} (b)). In our study, we obtain the ground-truth T-score by conducting tracking with a fixed tracker~\cite{li2019siamrpn++}. To be specific, we initialize the tracker with the ground-truth target region in a given frame, and conduct tracking in all remaining frames. With the fixed tracker and the almost identical tracking video (except the given template frame itself), only the template patch quality influences the tracking performance. Therefore, the obtained mean IoU reflects the desired template quality and is adopted as the ground-truth T-score.

\vspace{\abovesec}
\subsection{Score prediction}
\vspace{\belowsec}
\label{sec:arch}
We next introduce the proposed module for RT-scores prediction. In each frame, the module refers to the frame, query, and grounded box to generate the RT-score prediction. We re-use the fused feature from ``grounding'' as the per-frame visual-textual representation to boost the inference speed. 
As shown in Figure~\ref{fig:arch}, the proposed module takes the grounded region and the fused visual-textual feature from ``grounding''~\cite{yang2019fast} as inputs and predicts the scores for the grounded region. The module consists of three stand-alone $1\times 1$ convolutional layers. The RT-scores in the same spatial location as the top-1 ``grounding'' prediction is output as the final score prediction.

The score prediction module is trained separately from ``grounding'' and ``tracking.'' We model the R- and T-score predictions as two separate regression problems trained by the smoothed-L1 loss~\cite{girshick2015fast}. With a pre-trained grounding model~\cite{yang2019fast}, we generate training samples by collecting the triplets of visual-textual features, grounded regions, and derived RT-scores.
During training, we filter out the samples with a grounding confidence score of less than $0.5$. Such grounded regions are likely to be incorrect and can be well identified by grounding confidences. We find the filtering simplifies the score prediction problem and empirically leads to better performances. During inference, we consider such a region incorrect and directly set the R-score to $0$.

\vspace{\abovesec}
\subsection{Adaptive integration}
\vspace{\belowsec}
\label{sec:int}
In each frame, adaptive integration updates the tracking template and generates the final prediction based on the per-frame scores, instead of pre-defined rules or fixed weights.
With the RT-scores predicted, there exist multiple ways of generating the final prediction and updating the template based on the score, \eg, score-guided soft weighted fusion, or hard switching between grounding and tracking. We observe that the quality of the score prediction instead of the exact integration method influences the performance the most. Therefore, we present a vanilla version of hard switching as follows, and defer the introduction and experiments of the alternatives to Section~\ref{sec:ablation}. First, the R- and T-scores are multiplied in each frame to obtain a combined score that guides ``integration.'' We consider ``grounding'' more important whenever the predicted combined score is higher than the previously saved highest value. In such frames, we adopt ``grounding'' as the output and update the template. Otherwise, we output tracking predictions. 

With the same set of importance scores, we observe that the exact score-guided adaptive integration method, \eg, soft weighted fusion or hard switching, has no significant influence on the final performance. Instead, for ``integration,'' we find it important to accurately define and predict the importance scores based on the per-frame context. We detail the analyses and experiments in Section~\ref{sec:ablation}.

\begin{algorithm}[t]
\DontPrintSemicolon
\SetAlgoLined
\SetNoFillComment
\LinesNotNumbered 
\KwIn{Video $\mathcal{V}=\{v_1,\dots,v_n \}$ and Query $Q$}
\quad Function $\mathfrak{G}$ is the ``grounding'' module.\\
\quad Function $\mathfrak{T}$ is the ``tracking'' module.\\
\quad Function $\mathfrak{I}$ is the RT-score prediction module.\\
\quad $S$ is the saved score for the current template $T$.\\
\quad $b_g$ is the per-frame grounding prediction.\\
\quad $s_t$ is the RT-scores for the grounding prediction.\\
\quad $\lambda$ is the decay rate of the saved score $S$.\\
\KwOut{Per-frame object boxes $\mathcal{B}=\{b_1,\dots,b_n \}$}
\BlankLine
$b_g$ $\leftarrow$ $\mathfrak{G}$($v_1,Q$) \tcp*{Grounding}
$s_1$ $\leftarrow$ $\mathfrak{I}$($v_1,Q,b_g$) \tcp*{Initial RT-scores}
$b_1, S, T$ $\leftarrow$ $b_g, s_1, b_g$ \tcp*{Output, init tracker}
\For{$t$ in $2,\dots,n$}{
    $b_g$ $\leftarrow$ $\mathfrak{G}$($v_t,Q$)\\
    $s_t$ $\leftarrow$ $\mathfrak{I}$($v_t,Q,b_g$) \tcp*{Predicted RT-scores}
    \tcc{If grounding is more important}
    \uIf{$S<s_t$}{
    $b_t, S, T$ $\leftarrow$ $b_g, s_t, b_g$}
    \tcc{If tracking is more important}
    \Else{
    $b_t$ $\leftarrow$ $\mathfrak{T}$($v_t,T$)\\
    $S$ $\leftarrow$ $S*\lambda$
    }
}
\caption{{\bf Our implementation of GTI} \label{algo:inf}}
\end{algorithm}

\vspace{\abovesec}
\section{Implementation of GTI}
\vspace{\belowsec}
\label{sec:imp}
In this section, we present our real-time implementation of the GTI framework. We introduce the adopted ``grounding'' and ``tracking'' modules, as well as the overall pipeline.

\noindent\textbf{Grounding.}
Given a frame, the ``grounding'' module predicts a region based on the language query. We adopt the one-stage visual grounding~\cite{yang2019fast} as the grounding module because of its state-of-the-art accuracy and real-time inference speed. The grounding method merges language and spatial features into YOLOv3~\cite{redmon2018yolov3} for visual grounding. DarkNet-53~\cite{redmon2018yolov3} and feature pyramid network~\cite{lin2017feature} are used to encode the visual feature. With an input resolution of $256\times256$, the three feature pyramid heads have the spatial resolutions of $8\times8$, $16\times16$ and $32\times32$, respectively. Similar to one-stage object detection~\cite{redmon2018yolov3}, the grounding method outputs multiple box predictions at each of the $8\times 8+16\times 16+32\times 32=1344$ locations. With three anchor boxes predicted at each location, the method outputs $3\times1344=4032$ grounding predictions per frame. Each predicted region consists of five values, i.e. the relative position, width, height and the confidence score. The prediction with the highest confidence score is output as the final grounded region in each frame.

\noindent\textbf{Tracking.}
Given a frame, the ``tracking'' module localizes the target based on the language grounded region history in previous frames. We adopt the SiamRPN++~\cite{li2019siamrpn++} as the tracker in our implementation while various other object tracking methods~\cite{bhat2019learning,danelljan2019atom} can also be directly applied, as shown in ablation studies. SiamRPN++ is a Siamese network based tracker that models tracking as the feature cross-correlation between the tracking template and current frame.

\noindent\textbf{Inference.}
We then present the inference pipeline on a testing video in Algorithm~\ref{algo:inf}. Given no region history is available in the first frame, the ``grounding'' result is directly adopted as the output and used to initialize ``tracking.'' The predicted RT-scores are also saved. In all the following frames, the three modules operate simultaneously. ``Integration'' predicts the RT-scores in a frame and compare it to the saved value. Whenever a higher score appears, we adopt ``grounding'' as the output, and update tracking template $T$ and saved score $S$ accordingly. In remaining frames, ``tracking'' is adopted as the output. 

%% file: exp_new.tex
\vspace{\abovesec}
\section{Experiments}
\vspace{\belowsec}
\label{sec:exp}
\vspace{\abovesec}
\subsection{Datasets}
\vspace{\belowsec}
\begin{figure*}
    \centering
    \includegraphics[width=16cm]{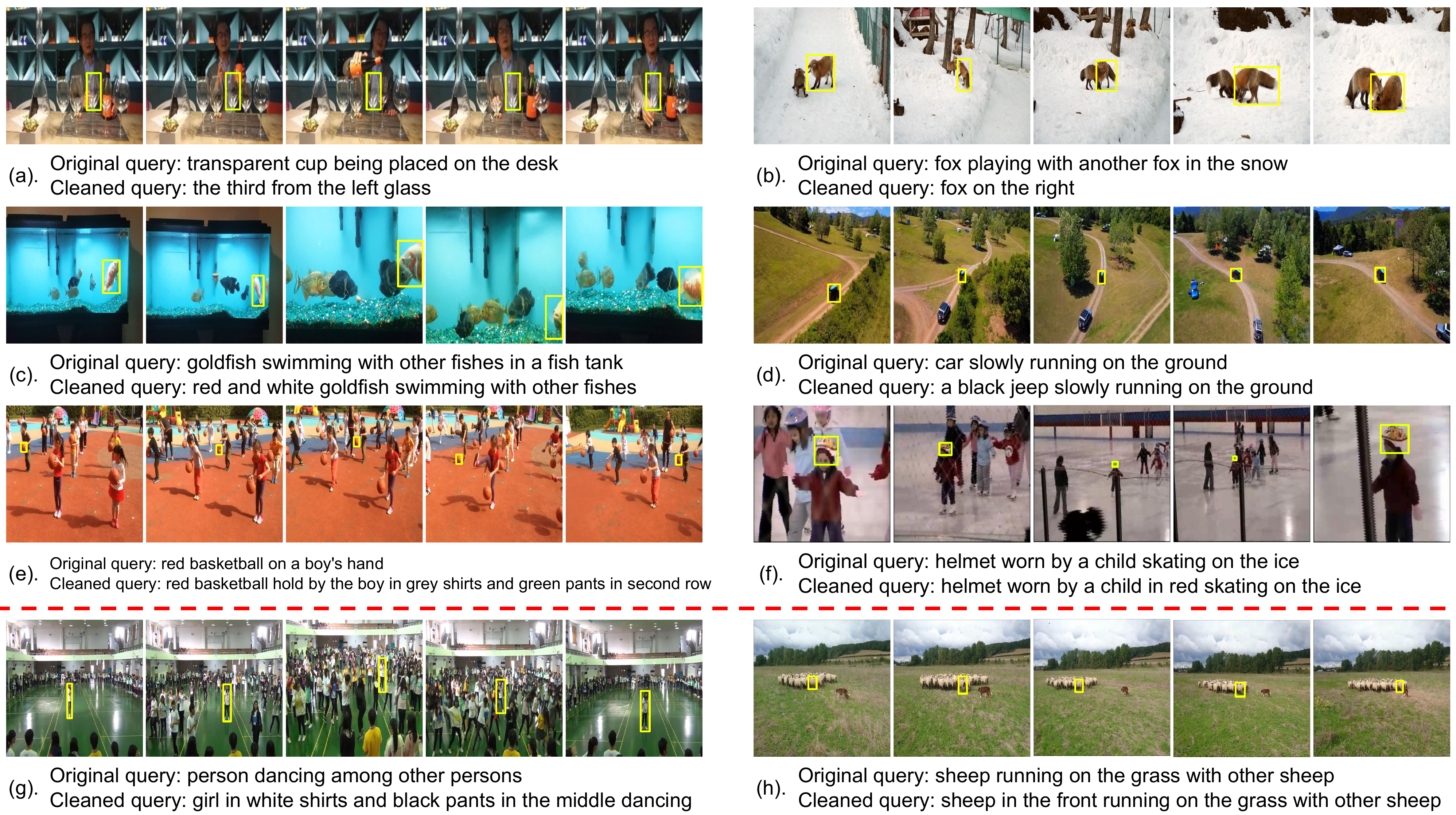}
    \caption{
    Examples of the disambiguated queries in LaSOT. The first three rows show the disambiguated queries, and the last row presents the samples that annotators find difficult to refer by language.
    }
    \vspace{-0.1in}
\label{fig:lasot}
\end{figure*}%
\noindent\textbf{Disambiguated LaSOT.} 
LaSOT~\cite{fan2019lasot} contains 1,400 videos with auxiliary language queries. We follow the split~\cite{fan2019lasot} that uses 1,120 videos for training and 280 videos for testing. The averaged video length is around 2,500 frames. 

The original LaSOT dataset~\cite{fan2019lasot} contains auxiliary language queries that might provide ambiguous target specifications. For example, in Figure~\ref{fig:lasot} (a), the referred glass can not be distinguished based on the original query. To facilitate tracking by language studies, we clean the LaSOT queries by replacing the ambiguous queries with new annotations. As the first step, annotators are presented with the video, target tubelet, and the original language query in LaSOT, and are asked to label if the target can be distinguished based on the original query. The collected annotations show that $322$ out of the $1,400$ original video queries are ambiguous. Annotators then generate new queries that have clear target specifications. Extra descriptions of the target's location, color, size, relationships are included in the cleaned queries. In the end, we verify the quality of the generated queries. Among the $322$ updated queries, $80$ queries are still ambiguous, \ie, at least one out of two annotators can not distinguish the target based on the new query. We fail to generate precise queries for all targets because some videos contain visually identical objects and are not proper for tracking by language studies (\eg, Figures~\ref{fig:lasot} (g) and (h)).

We provide representative examples of the updated queries in Figure~\ref{fig:lasot}. Figures~\ref{fig:lasot} (a) and (b) add extra location descriptions to disambiguate the query. Figures~\ref{fig:lasot} (c) and (d) include color and entity descriptions to provide the target specification. Figures~\ref{fig:lasot} (e) and (f) provide relationships and other detailed descriptions to generate a precise target specification. After the manual annotation, a small portion of samples is still ambiguous because the language query alone can not generate a clear specification for the given target. For example, in Figures~\ref{fig:lasot} (g) and (h), visually similar objects exist and make language referring difficult.

\noindent\textbf{Lingual OTB99.}
Lingual OTB99~\cite{li2017tracking} augments the OTB100 object tracking dataset~\cite{lu2014online,wu2013online} with natural language descriptions. One query is annotated per target object. We follow the training/ testing split~\cite{li2017tracking} that uses the OTB51 videos for training and the remaining 48 videos for testing. The averaged video length is around 600 frames. 

\noindent\textbf{Lingual ImageNet videos.}
The Lingual ImageNet videos dataset~\cite{li2017tracking} augments the ImageNet Video Object Detection dataset~\cite{russakovsky2015imagenet} with one query per target object. We follow the same split~\cite{li2017tracking} that uses $50$ videos for training and the other $50$ for testing. The averaged video length is around $270$ frames.

The targets and videos in the Lingual ImageNet videos dataset~\cite{li2017tracking} used in previous studies~\cite{li2017tracking} are far from real and oversimplify the problem, and thus are not suitable for study. We show the analyses in Section~\ref{sec:result}.

\vspace{\abovesec}
\subsection{Implementation details}
\vspace{\belowsec}
\noindent\textbf{Evaluation criteria.}
We evaluate the methods with precision and success scores~\cite{wu2015object}. The precision score reflects the percentage of frames where the estimated location falls within a given threshold of 20 pixels with the target. The success plot shows the ratio of success frames under an IoU threshold ranging from 0 to 1. The Area Under Curve (AUC) of the success plot represents the averaged success rates with different sampled thresholds and is used for evaluation. We follow the online tracking setting that the method only observes the previous and current frames for prediction.

\noindent\textbf{Training details.}
We train the score prediction module in RT-integration separately from the grounding and tracking modules. The three convolutional layers in the score prediction module have $D=512, 256, 6$ output channels, respectively. 
We train the model with RMSProp~\cite{tieleman2012lecture} and use a batch size of 32. The initial learning rate is $10^{-4}$ and follows a linear schedule. \eat{Equal weights are assigned to the R-score classification, regression and T-score regression.} We fine-tune the grounding module~\cite{yang2019fast} pre-trained on Flickr30K Entities~\cite{plummer2017flickr30k} with training set videos. For the tracking module, we use the models released by SiamRPN++~\cite{li2019siamrpn++} and fix the weights. The decay rate in Algorithm~\ref{algo:inf} is set to $0.998$.

\vspace{\abovesec}
\subsection{Experiment protocols}
\vspace{\belowsec}
\label{sec:setting}
 
Table~\ref{table:main} reports the tracking results on LaSOT~\cite{fan2019lasot} and Lingual OTB99~\cite{li2017tracking}. One-stage grounding~\cite{yang2019fast} is used for ``grounding'' and SiamRPN++~\cite{li2019siamrpn++} is used for ``tracking'' in all reported results expect the original LSAN~\cite{li2017tracking}. We list in the ``Integration guidance'' column the different integration methods. The \textbf{top portion} of Table~\ref{table:main} contains naive integration with either pre-defined scheduling rules or fixed fusion weights. Frame indexes such as ``all,'' ``first,'' and ``fixed interval'' indicate pre-defined scheduling is adopted and on which frames grounding is assigned higher importance. The \textbf{bottom portion} of the table contains the results of our adaptive integration methods. The types of adopted importance scores are listed in ``Integration guidance.''

\begin{table*}[t]
\centering
\caption{Tracking by language results on LaSOT~\cite{fan2019lasot} and Lingual OTB99~\cite{li2017tracking}.}
\begin{tabular}{ l l c c c c }
    \hline
    \multirow{2}{*}{Method} & Integration Guidance & \multicolumn{2}{c}{LaSOT} & \multicolumn{2}{c}{Lingual OTB99}\\
     & (\scriptsize{{\bf see Sec.~\ref{sec:setting} for detail}}) & Success & Precision & Success & Precision\\
    \hline
    \multicolumn{3}{l}{\textit{Single Module and Simple Combination Baselines}} & \\
    \hline
    Visual grounding & All & 0.416 & 0.411 & 0.442 & 0.551 \\
    First frame tracking & First & 0.331 & 0.301 & 0.421 & 0.551\\
    Middle frame tracking & Middle & 0.369 & 0.345 & 0.432 & 0.516 \\
    Last frame tracking & Last & 0.307 & 0.277 & 0.448 & 0.540 \\
    Random frame tracking\quad\quad & Random & 0.361 & 0.328 & 0.434 & 0.514 \\
    Fixed interval tracking & Fixed interval=5 & 0.423 & 0.420 & 0.449 & 0.552 \\ 
    Fixed interval tracking & Fixed interval=10 & 0.422 & 0.418 & 0.449 & 0.554 \\
    Fixed interval tracking & Fixed interval=20 & 0.420 & 0.412 & 0.449 & 0.556 \\
    LSAN~\cite{li2017tracking} & Fixed weights fusion & - & - & 0.259 & - \\
    LSAN++~\cite{li2017tracking} & Fixed weights fusion & 0.404 & 0.405 & 0.449 & 0.548 \\
    Feng~\etal~\cite{feng2020real} & Tracking by detection & 0.28 & 0.28 & 0.54 & \textbf{0.78} \\
    \hline
    \multicolumn{3}{l}{\textit{Different Variations of Our Methods}} & & \\
    \hline
    \textbf{Ours}-Grounding score & {Max grounding score} & 0.450 & 0.450 & 0.532 & 0.657 \\
    \textbf{Ours}-R score & {Max R-score} & 0.474 & 0.467 & 0.565 & 0.706 \\
    \textbf{Ours}-RT scores & {Max RT-scores} & \textbf{0.478} & \textbf{0.476} & \textbf{0.581} & 0.732 \\ 
    \hline
\end{tabular}
\vspace{-0.15in}
\label{table:main}
\end{table*}
Various baselines and state-of-the-art methods are experimented and compared. To be specific, we systematically study the following settings:
\begin{itemize} 
\item{\bf Visual grounding.}
One could attempt to approach tracking by language by processing each frame independently by grounding. One-stage visual grounding~\cite{yang2019fast} is adopted for the experiment.
\item{\bf First frame tracking.}
By taking the grounded region in the first frame as the tracking template, tracking by language is converted to a object tracking problem. This baseline is referred to as ``First frame tracking.''
\item{\bf Middle/ Last/ Random frame tracking.}
We initialize the tracker with the grounded region in the middle, last or one random sampled frame.
\item{\bf Fixed interval tracking.}
In this baseline, ``grounding'' is assigned a higher importance with a fixed temporal interval. We design the fixed interval to be similar to the averaged frequency of our adaptive integration.
\item{\bf LSAN/ LSAN++.}
We compare to the state-of-the-art tracking by language method LSAN~\cite{li2017tracking}. For a fair comparison, we strength LSAN with stronger grounding~\cite{yang2019fast} and tracking~\cite{li2019siamrpn++} backbones used in other experiments, and refer to it as ``LSAN++.''
\item{\bf Ours-Grounding/ R/ RT scores.}
We experiment with different variations of our methods.
\eat{We use the confidence score generated by ``grounding'' to guide the integration.}``Ours-'' indicates that the GTI implementation in Section~\ref{sec:imp} is adopted, with different importance score selections.
\end{itemize}

\vspace{\abovesec}
\subsection{Tracking by language results}
\label{sec:result}
\vspace{\belowsec}
\label{sec:results}
\noindent\textbf{Lingual OTB99.}
As the single module baseline, we first benchmark the ``grounding'' and ``tracking'' modules adopted in all following experiments. ``Grounding'' alone generates a success score of $\texttt{0.442}$, and ``tracking'' obtains a comparable success score of around $\texttt{0.434}$, as shown in ``Visual grounding'' and ``First/ middle/ last/ random frame tracking.''

``Integration'' aims to improve the tracking by language performance by synergistically combining the two modules. The top portion of Table~\ref{table:main} shows several simple combinations. Fixed temporal scheduling is one possible solution that switches between grounding and tracking with a fixed interval. ``Fixed interval tracking'' obtains a success score of $\texttt{0.449}$ and slightly outperforms the single module baseline. ``LSAN''~\cite{li2017tracking} fuses the two modules' predictions with a fixed weight applied in all frames. With the strengthened backbones, ``LSAN++'' generates a success score of $\texttt{0.449}$. In short, we observe limited improvements of less than $\texttt{0.01}$ over the single module baseline on all simple integration methods. The limited improvements confirm that the ``integration'' task is nontrivial, and that the {\it simple combination methods are ineffective}. 

Our \textit{first} contribution is proposing the new GTI framework, where we address ``integration'' as a score-guided self-judging process. The comparison between the top and bottom portions of Table~\ref{table:main} shows the importance of guiding integration with the scores predicted from the corresponding frame, language query, and box. One natural choice of the score is the grounding confidence. ``Ours-Grounding score'' reports a success score of $\texttt{0.532}$, which is significantly better than the grounding baseline ($\texttt{0.442}$) and the simple integration ($\texttt{0.449}$). The improvements show the advantage of score-guided integration instead of the simple combinations.

Our \textit{second} contribution is proposing better integration scores.\eat{With the proposed RT-integration, our method outperforms the ``Ours-Grounding score,'' as } As shown in the bottom portion of Table~\ref{table:main}, the ``Ours-R score'' achieves a success score of $\texttt{0.565}$, compared to  $\texttt{0.532}$ by ``Ours-Grounding score,'' $\texttt{0.449}$ by ``LSAN++,'' and  $\texttt{0.449}$ by ``fixed interval tracking.'' By jointly considering the template quality score, we further improve the success score. T-score alone does not work well because of the loss of region correctness information.

\noindent\textbf{LaSOT.}
``Grounding'' provides a baseline success score of $\texttt{0.416}$. The tracking baseline has a lower performance of $\texttt{0.361}$. ``Tracking'' performs relatively worse on LaSOT than OTB99 because the longer averaged video length in LaSOT makes tracking more challenging. For the same reason, updating the template multiple times performs better than a single template frame (cf. different intervals in ``Fixed interval tracking''). To eliminate the influence of the template update frequency, we design ``Fixed interval tracking'' to have a similar frequency as our RT-integration, which ranges from 5 to 20 frames. ``Ours-Grounding/ R/ RT score'' updates the template every $17.0/ 20.6/ 23.5$ frames on LaSOT and $7.9/ 13.8/ 16.6$ frames on Lingual OTB99. By eliminating the influence of the template update frequency, we show our adaptive integration performs better purely by more effective combining grounding and tracking.

We draw from LaSOT largely the same observation on ``integration'' as from Lingual OTB99. The simple integration methods such as ``LSAN++'' and ``fixed interval tracking'' show limited improvements over the single module baseline, while our adaptive integration significantly improves the performance. Our RT-integration achieves a success score of $\texttt{0.478}$, compared to $\texttt{0.404}$ by LSAN++ and  $\texttt{0.423}$ by fixed interval tracking. Clearly, the improvement shows the importance of score-guided integration and the effectiveness of our RT-integration.

The experiments in Table~\ref{table:main} are conducted on the disambiguated LaSOT dataset, except the original LSAN~\cite{li2017tracking} and Feng~\etal~\cite{feng2020real}. To examine the benefit of LaSOT query cleaning, we compare our full model ``Ours-RT score'' to the one trained on original LaSOT queries. ``Ours-RT score'' trained and tested with the original LaSOT queries generates a success score of $\texttt{0.475}$ and precision score of $\texttt{0.469}$. Compared to the performance of $\texttt{0.478}$ and $\texttt{0.476}$ after cleaning, the improvement is marginal. We expect the provided disambiguated queries open up the possibility of further improving tracking by language in future studies.

\noindent\textbf{Lingual ImageNet videos.}
We find that the Lingual ImageNet videos dataset is a special easy case, where current visual grounding methods already performs better than tracking by boxes (``Visual grounding'' success score:  $\texttt{0.864}$, ``SiamRPN++~\cite{li2019siamrpn++}'': $\texttt{0.768}]$). In Lingual ImageNet videos, the target objects are mostly in the center of the frame with few distracting objects exist, which makes the task easy for visual grounding. Despite the good results on this specific dataset, such videos are far from real and oversimplify the tracking by language problem.

\noindent\textbf{Inference speed.}
A fast inference speed is important for tracking by language. We evaluate the inference speed of our GTI implementation on a desktop with Intel Core i9-9900K@3.60GHz and NVIDIA 1080TI. Our framework runs at around \textbf{20 fps}, where the grounding module takes $20ms$ and the tracking module takes $30ms$. The proposed RT-integration module takes less than \textbf{1ms} by reusing the visual-textual features from grounding.

\begin{figure*}[t]
\begin{center}
  \centerline{\includegraphics[width=16cm]{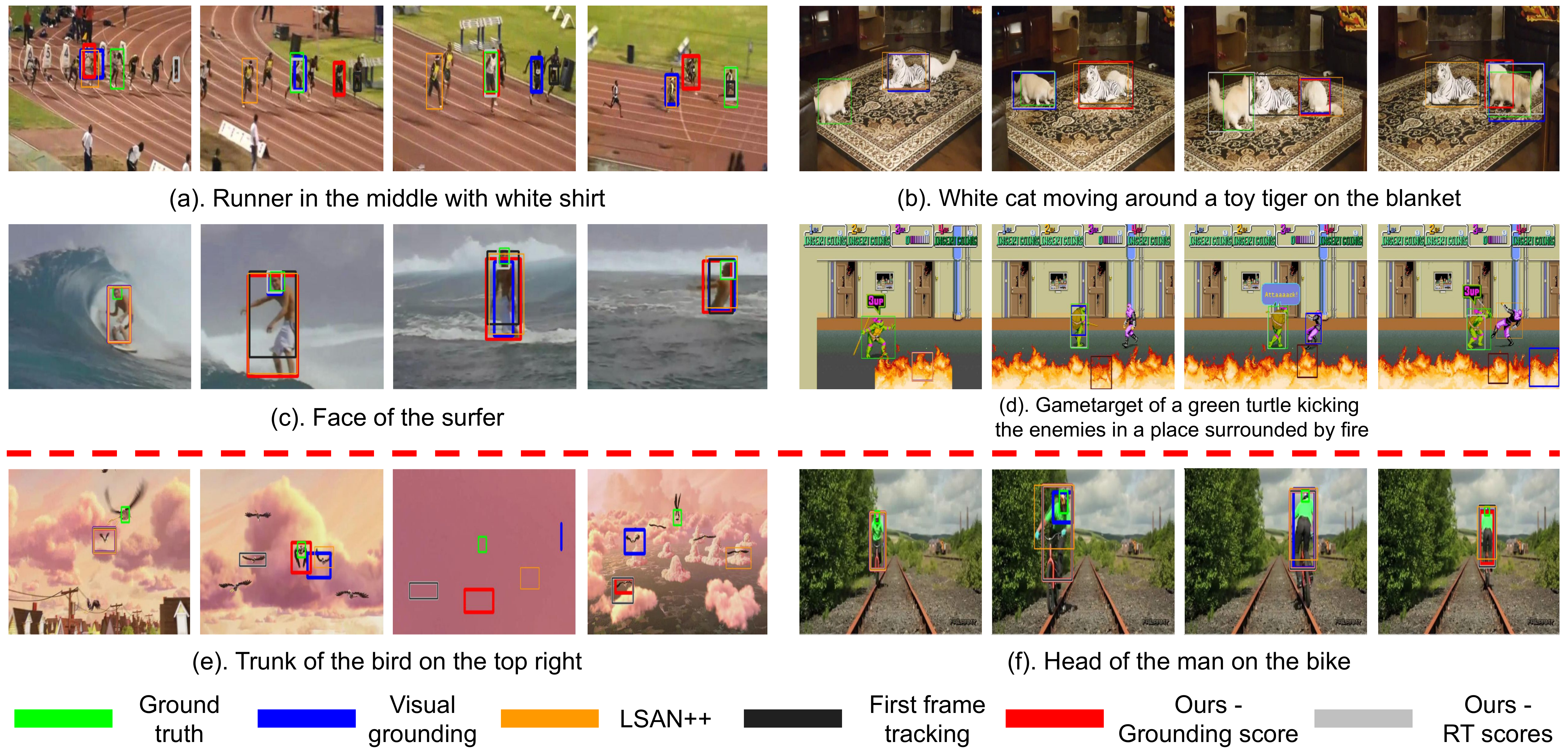}}
\end{center}
\vspace{-0.3in}
	\caption{Representative success cases (top two rows) and failures (bottom row) of our method. Figure (b), (d) are from LaSOT and the others are from Lingual OTB99. Best viewed in color and zoomed in.}
\label{fig:visu}
\end{figure*}
\begin{figure*}[t]
\begin{center}
  \centerline{\includegraphics[width=16cm]{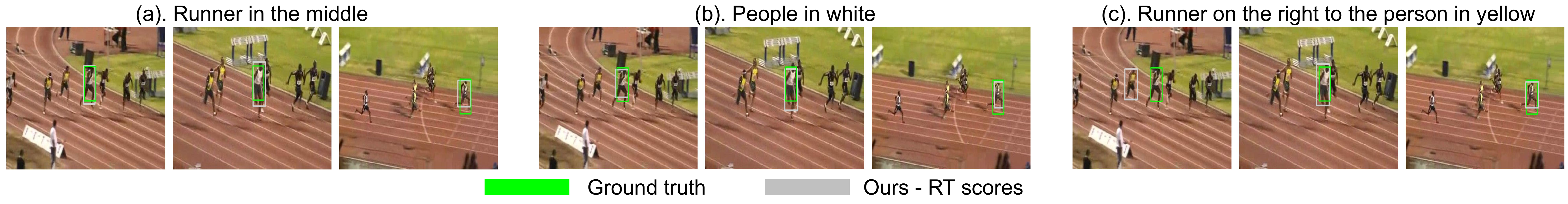}}
\end{center}
\vspace{-0.35in}
	\caption{Qualitative examples of referring to the same object with different language queries at inference.}
\label{fig:vari}
\vspace{-0.1in}
\end{figure*}

\vspace{\abovesec}
\subsection{Qualitative results analyses}
\vspace{\belowsec}
\label{sec:demo}
In this section, we compare the success and failure cases of the methods with naive integration modules as well as ours, to show the significance of our proposed RT-integration. We show representative examples in Figure~\ref{fig:visu}.
First, our method (silver boxes) are more stable and accurate when compared to per-frame visual grounding outputs (blue boxes). Including ``tracking'' (dark grey boxes) generates more stable results by exploiting the cross-frame visual similarity. However, the grounded region for tracker initialization in a randomly selected frame might be incorrect and thus fails ``tracking'' in the following frames. Figures~\ref{fig:visu} (a) and (b) show failure cases for the ``First frame tracking'' that our method can solve. \eat{``Ours - Grounding score'' guides integration with grounding confidences and thus generates much better results. Nonetheless, our proposed RT-scores are more effective in guiding integration.} Figures~\ref{fig:visu} (c) and (d) present challenging cases where ``grounding'' fails in most frames. When all compared methods fail, our RT-integration successfully combines grounding and tracking to provide mostly correct tracking results throughout the video.
Overall, our proposed approach performs better by more effectively integrating grounding with tracking.

    Despite the effectiveness of our proposed integration, when grounding fails on all frames, there is no hope to get correct results (cf. Figure~\ref{fig:visu} (e)). RT-score estimation may also be incorrect.  Figure~\ref{fig:visu} (f) shows an example that could be corrected, while our method fails to predict the correct RT-scores and correct the errors. Such failures are the cause of the gap from the oracles in Table~\ref{table:oracle}.

Furthermore, we experiment with referring to the same object with different language queries. We generate additional testing queries that describe different aspects of the target, \eg, color, location, or the relationship with other objects. Figure~\ref{fig:vari} shows good qualitative results that the method generalizes well onto free-form referring queries.
\vspace{\abovesec}
\subsection{Oracle analyses}
\label{sec:oracle}
\vspace{\belowsec}

\begin{table}[t]
\centering
\caption{Oracle analyses of tracking by language or gt boxes.}
\begin{tabular}{ l l c c c c }
    \hline
    \multirow{2}{*}{Method} & Object & \multicolumn{2}{c}{LaSOT} & \multicolumn{2}{c}{Lingual OTB99}\\
     & referring & Succ. & Prec. & Succ. & Prec.\\
    \hline
    MEEM~\cite{zhang2014meem} & GT box & 0.257 & 0.227 & 0.491 & 0.725 \\
    HCFT~\cite{ma2015hierarchical} & GT box & 0.250 & 0.241 & 0.518 & 0.778 \\
    ECO~\cite{danelljan2017eco} & GT box & 0.324 & 0.301 & 0.675 & 0.888 \\
    SiamFC~\cite{bertinetto2016fully} & GT box & 0.336 & 0.339 & - & - \\
    VITAL~\cite{song2018vital} & GT box & 0.390 & 0.360 & - & - \\
    MDNet~\cite{nam2016learning} & GT box & 0.397 & 0.373 & - & - \\
    SiamRPN~\cite{li2018high} & GT box & 0.449 & 0.435 & 0.687 & 0.897 \\
    \footnotesize{SiamRPN++~\cite{li2019siamrpn++}} & GT box & 0.496 & 0.489 & \textbf{0.698} & \textbf{0.899} \\
    \makecell[l]{Ours-R-oracle}  & \footnotesize{Language} & 0.624 & 0.656 & 0.645 & 0.826\\
    \makecell[l]{Ours-RT-oracle} & \footnotesize{Language} & \textbf{0.631} & \textbf{0.665} & 0.672 & 0.863 \\
    \hline
\end{tabular}
\label{table:oracle}
\end{table}

Tracking by language is generally more challenging than the conventional tracking by gt box setting and tends to perform worse on the same video~\cite{li2017tracking}. The proposed RT-integration greatly improves the tracking by language performance. However, the score prediction module meanwhile introduces new errors and potentially limits the overall performance. In this section, we examine the upper bound of the GTI framework that has an ideal ``integration'' model, given the status quo of grounding~\cite{yang2019fast} and tracking~\cite{li2019siamrpn++}. We compare the oracles to both tracking by language and by gt box results, specifically the following settings:
\begin{itemize}
\item{\bf Tracking by gt box.}
With the same dataset split, tracking by ground-truth box~\cite{danelljan2017eco,bertinetto2016fully,song2018vital,nam2016learning,li2019siamrpn++,li2018high} serves as an upper bound of tracking with ideal target specifications.
\item{\bf Ours-R-oracle.}
We design two oracle analyses with the same GTI implementation in Section~\ref{sec:imp}. The R-score in the oracle analyses is calculated with the ground-truth box at each frame instead of predicted. 
\item{\bf Ours-RT-oracle.}
``Ours-RT-oracle'' considers both the region correctness and template quality scores. 
\end{itemize}
As shown in Table~\ref{table:oracle}, the GTI framework with an ideal integration module achieves comparable (on shorter videos~\cite{li2017tracking}) if not better (on longer videos~\cite{fan2019lasot}) performance than the state-of-the-art tracker~\cite{li2019siamrpn++}. The good oracle performance implies the possibility of tracking by language to achieve comparable results to tracking by gt box, despite the more challenge setting. Meanwhile, the existing gap between the performance of the oracle GTI and our implementation shows that the integration problem is non-trivial, and motivate us to develop better integration methods in future studies. Finally, with the continuously improving grounding and tracking methods, we expect the future GTI frameworks with stronger modules to further improve the tracking by language performances.

\subsection{Ablation study}
\label{sec:ablation}
In this section, we conduct ablation studies to understand our method better. We first compare alternative adaptive integration methods to the hard switch approach introduced in Section~\ref{sec:int}. We then show that our proposed ``integration'' module's importance and effectiveness hold under different ``grounding'' and ``tracking'' backbones.

\noindent\textbf{Adaptive integration.}
Given the obtained scores for integration, there are alternative methods to the hard switch approach described in Section~\ref{sec:int}. We experiment with other adaptive integration methods to examine their influences on the performance. We conduct the ablation studies on adaptive integration with the oracle RT-scores detailed in Section~\ref{sec:oracle} to eliminate the influence of score prediction quality. 

Given the predicted tracking results, grounding results, and integration scores, the final step is to integrate the tracking and grounding prediction with the predicted scores for both tracking template update and current frame prediction. We first explore three alternative ways of updating the tracking template.
Our adopted ``\texttt{greedy update}'' option outputs the grounding prediction and updates the tracking template whenever a higher score appears. ``\texttt{Improvement threshold}'' follows the same greedy update protocol with a tuned score improvement threshold of $20\%$ included. ``\texttt{Fixed weight update}'' consists of a memory module and updates the template's visual feature with a fixed update rate of $0.9$~\cite{yang2018learning}. ``\texttt{Score weighted update}'' further adopts the predicted scores as the update rate. 

Table~\ref{table:ablation} shows the success and precision scores of the compared adaptive integration methods. Our adopted approach generates a success score of $\texttt{0.672}$, which is comparable to the best score of $\texttt{0.675}$. We observe no significant gain by more complex alternatives and thus choose the simple yet effective ``\texttt{greedy update}'' as the adopted approach.

Other than the tracking template update, the adaptive integration method also generates the final prediction at each frame. Our adopted ``hard switch'' method that outputs either tracking or grounding results based on the integration scores. As an alternative, we experiment with the ``\texttt{soft fusion}'' used by previous studies~\cite{li2017tracking}, where the output fusion is computed as a weighted sum of the grounding and tracking heatmaps with the predicted per-frame integration score. We observe the ``hard switch'' outperforms the ``soft fusion'' and thus adopt the ``hard switch'' approach for output fusion.

\begin{table}[t]
\centering
\caption{Tracking by language results with different adaptive integration methods on Lingual OTB99. The ``greedy update'' and ``hard switch'' result shown in the first row is the approach we adopted. We highlight the best the second-highest scores by \textbf{bold} and \underline{underline}, respectively.}
\begin{tabular}{ l c c c c c c }
    \hline
    Template update & Output fusion & Succ. & Prec. \\
    \hline
    \rowcolor{LightCyan}
    Greedy update & Hard switch & \underline{0.672} & \underline{0.863} \\
    Greedy update & Soft fusion & 0.646 & 0.843 \\
    Improvement threshold & Hard switch & 0.632 & 0.814 \\
    Fixed weight update & Hard switch & \textbf{0.675} & \textbf{0.867} \\
    Score weighted update & Hard switch & 0.668 & 0.856 \\
    \hline
\end{tabular}
\label{table:ablation}
\end{table}

\noindent\textbf{GTI backbones.}
We then experiment with the influence of ``integration'' with different ``grounding'' and ``tracking'' backbones. We replace the backbones with relatively weaker (but faster) modules and benchmark the corresponding GTI implementations. We replace the adopted SiamRPN++~\cite{li2019siamrpn++} with SiamRPN~\cite{li2018high}, and one-stage visual grounding~\cite{yang2019fast} with a lighter version Onestage-light~\cite{yang2019fast}. 

Table~\ref{table:module} shows the obtained results. In short, better grounding and tracking modules generally lead to better tracking by language performances. More importantly, our proposed ``RT-Integration'' brings significant success score improvements with different backbones (\cf~``Fixed interval tracking '' and ``Ours-RT scores'' with the same backbone). The consistent improvements of $0.179$, $0.109$, $0.132$ over the simple combination baseline indicate that ``RT-integration'' is effective under different grounding and tracking backbones. With the continuously improving grounding and tracking methods, we expect future GTI implementations to further improve the tracking by language performance.

\begin{table}[t]
\centering
\caption{Tracking by language results with different grounding and tracking backbones on Lingual OTB99. Comparing different integration methods with the same grounding and tracking module, the effectiveness of the proposed ``Ours-RT Scores'' holds under different backbones.}
\begin{tabular}{ l l l c c }
    \hline
    \multirow{2}{*}{Integration method} & \multirow{2}{*}{Grounding} & \multirow{2}{*}{Tracking} & \multicolumn{2}{c}{Lingual OTB99}\\
       &    &   & Succ. & Prec.\\
    \hline
    Visual Grounding & Onestage-light & None & 0.379 & 0.491 \\
    Visual Grounding & Onestage & None & 0.442 & 0.551 \\
    Fixed interval tracking & Onestage-light & SiamRPN++ & 0.391 & 0.492 \\
    Fixed interval tracking & Onestage & SiamRPN & 0.446 & 0.553 \\
    Fixed interval tracking & Onestage & SiamRPN++ & 0.449 & 0.554 \\
    \hline
    Ours-RT scores & Onestage-light & SiamRPN++ & 0.570 & 0.723 \\
    Ours-RT scores & Onestage & SiamRPN & 0.555 & 0.701 \\
    Ours-RT scores & Onestage & SiamRPN++ & 0.581 & 0.732 \\
    \hline
\end{tabular}
\label{table:module}
\end{table}

%% file: conclusion.tex
\vspace{\abovesec}
\vspace{-2pt}
\section{Conclusion}
\vspace{-1pt}
\vspace{\belowsec}

We have proposed a new GTI framework for tracking by language where we decompose the task into three sub-tasks: grounding, tracking, and integration. We focus on the key sub-task of ``integration'' that synergistically combines grounding and tracking, and propose an RT-integration module that defines two scores to guide integration in each frame. The R-score represents the region correctness, and the T-score represents the template quality. We benchmark our real-time implementation of the GTI framework on LaSOT and Lingual OTB99 to demonstrate highly promising results.

%% file: bare_jrnl.bbl
\begin{thebibliography}{10}
\providecommand{\url}[1]{#1}
\csname url@samestyle\endcsname
\providecommand{\newblock}{\relax}
\providecommand{\bibinfo}[2]{#2}
\providecommand{\BIBentrySTDinterwordspacing}{\spaceskip=0pt\relax}
\providecommand{\BIBentryALTinterwordstretchfactor}{4}
\providecommand{\BIBentryALTinterwordspacing}{\spaceskip=\fontdimen2\font plus
\BIBentryALTinterwordstretchfactor\fontdimen3\font minus
  \fontdimen4\font\relax}
\providecommand{\BIBforeignlanguage}[2]{{%
\expandafter\ifx\csname l@#1\endcsname\relax
\typeout{** WARNING: IEEEtran.bst: No hyphenation pattern has been}%
\typeout{** loaded for the language `#1'. Using the pattern for}%
\typeout{** the default language instead.}%
\else
\language=\csname l@#1\endcsname
\fi
#2}}
\providecommand{\BIBdecl}{\relax}
\BIBdecl

\bibitem{li2017tracking}
Z.~Li, R.~Tao, E.~Gavves, C.~G. Snoek, and A.~W. Smeulders, ``Tracking by
  natural language specification,'' in \emph{Proceedings of the IEEE Conference
  on Computer Vision and Pattern Recognition}, 2017, pp. 6495--6503.

\bibitem{VOT_TPAMI}
M.~Kristan, J.~Matas, A.~Leonardis, T.~Vojir, R.~Pflugfelder, G.~Fernandez,
  G.~Nebehay, F.~Porikli, and L.~\v{C}ehovin, ``A novel performance evaluation
  methodology for single-target trackers,'' \emph{IEEE Transactions on Pattern
  Analysis and Machine Intelligence}, vol.~38, no.~11, pp. 2137--2155, Nov
  2016.

\bibitem{wu2013online}
Y.~Wu, J.~Lim, and M.-H. Yang, ``Online object tracking: A benchmark,'' in
  \emph{Proceedings of the IEEE conference on computer vision and pattern
  recognition}, 2013, pp. 2411--2418.

\bibitem{wu2015object}
Y.~Wu, J.~Lim, and M.~Yang, ``Object tracking benchmark,'' \emph{IEEE
  Transactions on Pattern Analysis and Machine Intelligence}, vol.~37, no.~9,
  pp. 1834--1848, 2015.

\bibitem{yilmaz2006object}
A.~Yilmaz, O.~Javed, and M.~Shah, ``Object tracking: A survey,'' \emph{Acm
  computing surveys (CSUR)}, vol.~38, no.~4, p.~13, 2006.

\bibitem{yamaguchi2017spatio}
M.~Yamaguchi, K.~Saito, Y.~Ushiku, and T.~Harada, ``Spatio-temporal person
  retrieval via natural language queries,'' in \emph{ICCV}, 2017, pp.
  1453--1462.

\bibitem{lei2018tvqa}
J.~Lei, L.~Yu, M.~Bansal, and T.~L. Berg, ``Tvqa: Localized, compositional
  video question answering,'' in \emph{Empirical Methods in Natural Language
  Processing}, 2018.

\bibitem{plummer2017flickr30k}
B.~A. Plummer, L.~Wang, C.~M. Cervantes, J.~C. Caicedo, J.~Hockenmaier, and
  S.~Lazebnik, ``Flickr30k entities: Collecting region-to-phrase
  correspondences for richer image-to-sentence models,'' \emph{International
  journal of computer vision}, vol. 123, no.~1, pp. 74--93, 2017.

\bibitem{yang2019fast}
Z.~Yang, B.~Gong, L.~Wang, W.~Huang, D.~Yu, and J.~Luo, ``A fast and accurate
  one-stage approach to visual grounding,'' in \emph{ICCV}, 2019.

\bibitem{yu2018mattnet}
L.~Yu, Z.~Lin, X.~Shen, J.~Yang, X.~Lu, M.~Bansal, and T.~L. Berg, ``Mattnet:
  Modular attention network for referring expression comprehension,'' in
  \emph{Proceedings of the IEEE Conference on Computer Vision and Pattern
  Recognition}, 2018, pp. 1307--1315.

\bibitem{yu2016modeling}
L.~Yu, P.~Poirson, S.~Yang, A.~C. Berg, and T.~L. Berg, ``Modeling context in
  referring expressions,'' in \emph{European Conference on Computer
  Vision}.\hskip 1em plus 0.5em minus 0.4em\relax Springer, 2016, pp. 69--85.

\bibitem{danelljan2017eco}
M.~Danelljan, G.~Bhat, F.~Shahbaz~Khan, and M.~Felsberg, ``Eco: Efficient
  convolution operators for tracking,'' in \emph{CVPR}, 2017, pp. 6638--6646.

\bibitem{li2019siamrpn++}
B.~Li, W.~Wu, Q.~Wang, F.~Zhang, J.~Xing, and J.~Yan, ``Siamrpn++: Evolution of
  siamese visual tracking with very deep networks,'' in \emph{Proceedings of
  the IEEE Conference on Computer Vision and Pattern Recognition}, 2019, pp.
  4282--4291.

\bibitem{li2018high}
B.~Li, J.~Yan, W.~Wu, Z.~Zhu, and X.~Hu, ``High performance visual tracking
  with siamese region proposal network,'' in \emph{Proceedings of the IEEE
  Conference on Computer Vision and Pattern Recognition}, 2018, pp. 8971--8980.

\bibitem{fan2019lasot}
H.~Fan, L.~Lin, F.~Yang, P.~Chu, G.~Deng, S.~Yu, H.~Bai, Y.~Xu, C.~Liao, and
  H.~Ling, ``Lasot: A high-quality benchmark for large-scale single object
  tracking,'' in \emph{CVPR}, 2019, pp. 5374--5383.

\bibitem{danelljan2015learning}
M.~Danelljan, G.~Hager, F.~Shahbaz~Khan, and M.~Felsberg, ``Learning spatially
  regularized correlation filters for visual tracking,'' in \emph{Proceedings
  of the IEEE international conference on computer vision}, 2015, pp.
  4310--4318.

\bibitem{danelljan2014adaptive}
M.~Danelljan, F.~Shahbaz~Khan, M.~Felsberg, and J.~Van~de Weijer, ``Adaptive
  color attributes for real-time visual tracking,'' in \emph{Proceedings of the
  IEEE Conference on Computer Vision and Pattern Recognition}, 2014, pp.
  1090--1097.

\bibitem{henriques2014high}
J.~F. Henriques, R.~Caseiro, P.~Martins, and J.~Batista, ``High-speed tracking
  with kernelized correlation filters,'' \emph{IEEE transactions on pattern
  analysis and machine intelligence}, vol.~37, no.~3, pp. 583--596, 2014.

\bibitem{bertinetto2016fully}
L.~Bertinetto, J.~Valmadre, J.~F. Henriques, A.~Vedaldi, and P.~H. Torr,
  ``Fully-convolutional siamese networks for object tracking,'' in
  \emph{European conference on computer vision}.\hskip 1em plus 0.5em minus
  0.4em\relax Springer, 2016, pp. 850--865.

\bibitem{shi2019not}
J.~Shi, J.~Xu, B.~Gong, and C.~Xu, ``Not all frames are equal:
  Weakly-supervised video grounding with contextual similarity and visual
  clustering losses,'' in \emph{Proceedings of the IEEE Conference on Computer
  Vision and Pattern Recognition}, 2019, pp. 10\,444--10\,452.

\bibitem{wang2018describe}
X.~Wang, C.~Li, R.~Yang, T.~Zhang, J.~Tang, and B.~Luo, ``Describe and attend
  to track: Learning natural language guided structural representation and
  visual attention for object tracking,'' \emph{arXiv preprint
  arXiv:1811.10014}, 2018.

\bibitem{feng2020real}
Q.~Feng, V.~Ablavsky, Q.~Bai, G.~Li, and S.~Sclaroff, ``Real-time visual object
  tracking with natural language description,'' in \emph{The IEEE Winter
  Conference on Applications of Computer Vision}, 2020, pp. 700--709.

\bibitem{kazemzadeh2014referitgame}
S.~Kazemzadeh, V.~Ordonez, M.~Matten, and T.~Berg, ``Referitgame: Referring to
  objects in photographs of natural scenes,'' in \emph{Proceedings of the 2014
  conference on empirical methods in natural language processing (EMNLP)},
  2014, pp. 787--798.

\bibitem{plummer2018conditional}
B.~A. Plummer, P.~Kordas, M.~Hadi~Kiapour, S.~Zheng, R.~Piramuthu, and
  S.~Lazebnik, ``Conditional image-text embedding networks,'' in \emph{ECCV},
  2018, pp. 249--264.

\bibitem{sadhu2019zero}
A.~Sadhu, K.~Chen, and R.~Nevatia, ``Zero-shot grounding of objects from
  natural language queries,'' in \emph{Proceedings of the IEEE International
  Conference on Computer Vision}, 2019, pp. 4694--4703.

\bibitem{wang2019learning}
L.~Wang, Y.~Li, J.~Huang, and S.~Lazebnik, ``Learning two-branch neural
  networks for image-text matching tasks,'' \emph{IEEE Transactions on Pattern
  Analysis and Machine Intelligence}, vol.~41, no.~2, pp. 394--407, 2018.

\bibitem{yu2017joint}
L.~Yu, H.~Tan, M.~Bansal, and T.~L. Berg, ``A joint speaker-listener-reinforcer
  model for referring expressions,'' in \emph{CVPR}, 2017, pp. 7282--7290.

\bibitem{chen2018real}
X.~Chen, L.~Ma, J.~Chen, Z.~Jie, W.~Liu, and J.~Luo, ``Real-time referring
  expression comprehension by single-stage grounding network,'' \emph{arXiv
  preprint arXiv:1812.03426}, 2018.

\bibitem{yang2020improving}
Z.~Yang, T.~Chen, L.~Wang, and J.~Luo, ``Improving one-stage visual grounding
  by recursive sub-query construction,'' in \emph{ECCV}, 2020.

\bibitem{zhang2020does}
Z.~Zhang, Z.~Zhao, Y.~Zhao, Q.~Wang, H.~Liu, and L.~Gao, ``Where does it exist:
  Spatio-temporal video grounding for multi-form sentences,'' in \emph{CVPR},
  2020, pp. 10\,668--10\,677.

\bibitem{sadhu2020video}
A.~Sadhu, K.~Chen, and R.~Nevatia, ``Video object grounding using semantic
  roles in language description,'' in \emph{CVPR}, 2020, pp. 10\,417--10\,427.

\bibitem{danelljan2019atom}
M.~Danelljan, G.~Bhat, F.~S. Khan, and M.~Felsberg, ``Atom: Accurate tracking
  by overlap maximization,'' in \emph{Proceedings of the IEEE Conference on
  Computer Vision and Pattern Recognition}, 2019, pp. 4660--4669.

\bibitem{jiang2018acquisition}
B.~Jiang, R.~Luo, J.~Mao, T.~Xiao, and Y.~Jiang, ``Acquisition of localization
  confidence for accurate object detection,'' in \emph{Proceedings of the
  European Conference on Computer Vision (ECCV)}, 2018, pp. 784--799.

\bibitem{goldman2019precise}
E.~Goldman, R.~Herzig, A.~Eisenschtat, J.~Goldberger, and T.~Hassner, ``Precise
  detection in densely packed scenes,'' in \emph{Proceedings of the IEEE
  Conference on Computer Vision and Pattern Recognition}, 2019, pp. 5227--5236.

\bibitem{huang2019mask}
Z.~Huang, L.~Huang, Y.~Gong, C.~Huang, and X.~Wang, ``Mask scoring r-cnn,'' in
  \emph{Proceedings of the IEEE Conference on Computer Vision and Pattern
  Recognition}, 2019, pp. 6409--6418.

\bibitem{lin2017feature}
T.-Y. Lin, P.~Doll{\'a}r, R.~Girshick, K.~He, B.~Hariharan, and S.~Belongie,
  ``Feature pyramid networks for object detection,'' in \emph{Proceedings of
  the IEEE conference on computer vision and pattern recognition}, 2017, pp.
  2117--2125.

\bibitem{kang2016object}
K.~Kang, W.~Ouyang, H.~Li, and X.~Wang, ``Object detection from video tubelets
  with convolutional neural networks,'' in \emph{CVPR}, 2016, pp. 817--825.

\bibitem{tang2019object}
P.~Tang, C.~Wang, X.~Wang, W.~Liu, W.~Zeng, and J.~Wang, ``Object detection in
  videos by high quality object linking,'' \emph{IEEE TPAMI}, vol.~42, no.~5,
  pp. 1272--1278, 2019.

\bibitem{girshick2015fast}
R.~Girshick, ``Fast r-cnn,'' in \emph{Proceedings of the IEEE international
  conference on computer vision}, 2015, pp. 1440--1448.

\bibitem{redmon2018yolov3}
J.~Redmon and A.~Farhadi, ``Yolov3: An incremental improvement,'' \emph{arXiv
  preprint arXiv:1804.02767}, 2018.

\bibitem{bhat2019learning}
G.~Bhat, M.~Danelljan, L.~V. Gool, and R.~Timofte, ``Learning discriminative
  model prediction for tracking,'' in \emph{Proceedings of the IEEE
  International Conference on Computer Vision}, 2019, pp. 6182--6191.

\bibitem{lu2014online}
Y.~Lu, T.~Wu, and S.~Chun~Zhu, ``Online object tracking, learning and parsing
  with and-or graphs,'' in \emph{Proceedings of the IEEE Conference on Computer
  Vision and Pattern Recognition}, 2014, pp. 3462--3469.

\bibitem{russakovsky2015imagenet}
O.~Russakovsky, J.~Deng, H.~Su, J.~Krause, S.~Satheesh, S.~Ma, Z.~Huang,
  A.~Karpathy, A.~Khosla, M.~Bernstein \emph{et~al.}, ``Imagenet large scale
  visual recognition challenge,'' \emph{International journal of computer
  vision}, vol. 115, no.~3, pp. 211--252, 2015.

\bibitem{tieleman2012lecture}
T.~Tieleman and G.~Hinton, ``Lecture 6.5-rmsprop: Divide the gradient by a
  running average of its recent magnitude,'' \emph{COURSERA: Neural networks
  for machine learning}, vol.~4, no.~2, pp. 26--31, 2012.

\bibitem{zhang2014meem}
J.~Zhang, S.~Ma, and S.~Sclaroff, ``Meem: robust tracking via multiple experts
  using entropy minimization,'' in \emph{European conference on computer
  vision}.\hskip 1em plus 0.5em minus 0.4em\relax Springer, 2014, pp. 188--203.

\bibitem{ma2015hierarchical}
C.~Ma, J.-B. Huang, X.~Yang, and M.-H. Yang, ``Hierarchical convolutional
  features for visual tracking,'' in \emph{ICCV}, 2015, pp. 3074--3082.

\bibitem{song2018vital}
Y.~Song, C.~Ma, X.~Wu, L.~Gong, L.~Bao, W.~Zuo, C.~Shen, R.~W. Lau, and M.-H.
  Yang, ``Vital: Visual tracking via adversarial learning,'' in
  \emph{Proceedings of the IEEE conference on computer vision and pattern
  recognition}, 2018, pp. 8990--8999.

\bibitem{nam2016learning}
H.~Nam and B.~Han, ``Learning multi-domain convolutional neural networks for
  visual tracking,'' in \emph{Proceedings of the IEEE conference on computer
  vision and pattern recognition}, 2016, pp. 4293--4302.

\bibitem{yang2018learning}
T.~Yang and A.~B. Chan, ``Learning dynamic memory networks for object
  tracking,'' in \emph{Proceedings of the European Conference on Computer
  Vision (ECCV)}, 2018, pp. 152--167.

\end{thebibliography}
